\documentclass[11pt]{article}

\usepackage{amsmath,amssymb}
\usepackage{changepage}
\usepackage[utf8x]{inputenc}
\usepackage{cite}
\usepackage{nameref,hyperref}
\usepackage{microtype}
\usepackage[table]{xcolor}
\usepackage{array}
\usepackage{authblk}
\usepackage{lastpage,fancyhdr,graphicx}
\usepackage[margin=0.70in]{geometry}
\usepackage{amsmath}
\usepackage{graphicx}
\usepackage{mathtools}
\usepackage{cuted}
\usepackage{wrapfig}
\usepackage{xcolor}

\newcommand{\alg}{{PMCE}} 

\usepackage{balance}

\usepackage[aboveskip=1pt,labelfont=bf,labelsep=period,justification=raggedright,singlelinecheck=off]{caption}

\begin{document}

\title{\alg: efficient inference of expressive models of cancer evolution with high prognostic power}

\author[1]{Fabrizio Angaroni}
\author[2]{Kevin Chen}
\author[3,4]{Chiara Damiani}
\author[5]{Giulio Caravagna}
\author[6,7]{Alex Graudenzi}
\author[2,8,9]{Daniele Ramazzotti}
\affil[1]{Dept. of Informatics, Systems and Communication, Univ. of Milan-Bicocca, Milan, Italy}
\affil[2]{Dept. of Computer Science, Stanford University, USA}
\affil[3]{Dept. of Biotechnology and Biosciences, Univ. of Milan-Bicocca, Milan, Italy}
\affil[4]{Sysbio Centre for Systems Biology, Milan, Italy}
\affil[5]{Dept. of Mathematics and Geosciences, Univ. of Trieste, Trieste, Italy}
\affil[6]{Inst. of Molecular Bioimaging and Physiology, Consiglio Nazionale delle Ricerche (IBFM-CNR), Segrate, Milan, Italy}
\affil[7]{Bicocca Bioinformatics, Biostatistics and Bioimaging Centre (B4), Milan, Italy}
\affil[8]{Dept. of Pathology, Stanford University, USA}
\affil[9]{Dept. of Medicine and Surgery, Univ. of Milan-Bicocca, Monza, Italy}
\date{}

\maketitle 

\section*{Abstract}
\textbf{Motivation:} 
Driver (epi)genomic alterations underlie the positive selection of cancer subpopulations, which promotes drug resistance and relapse. 
Even though substantial heterogeneity is witnessed in most cancer types, mutation accumulation patterns can be regularly found and can be exploited to reconstruct predictive models of cancer evolution. Yet, available methods cannot infer logical formulas connecting events to represent alternative evolutionary routes or convergent evolution.  \\
\textbf{Results:}
We introduce \alg{}, an expressive framework that leverages mutational profiles from cross-sectional sequencing data to infer probabilistic graphical models of cancer evolution including arbitrary logical formulas, and which outperforms the state-of-the-art in terms of accuracy and robustness to noise, on simulations.  
\\
The application of \alg{} to $7866$ samples from the TCGA database allows us to identify a highly significant correlation between the predicted evolutionary paths and the overall survival in $7$ tumor types, proving that our approach can effectively stratify cancer patients in reliable risk groups.  
\\
\textbf{Availability:}  \alg{} is freely available at \href{https://github.com/BIMIB-DISCo/PMCE}{https://github.com/BIMIB-DISCo/PMCE}, in addition to the code to replicate all the analyses presented in the manuscript. \\
\textbf{Contacts:} \href{daniele.ramazzotti@unimib.it}{daniele.ramazzotti@unimib.it}, \href{alex.graudenzi@ibfm.cnr.it}{alex.graudenzi@ibfm.cnr.it}.

\section{Introduction}
Many natural phenomena are characterized by the presence of ordered sequences of discrete states or events, such as the accumulation of somatic mutations during cancer progression  \cite{caravagna2016algorithmic,10.1371/journal.pcbi.1007246}. 
A particular class of mathematical models used to represent such phenomena is provided by Bayesian Networks (BNs) and related extensions, such as Conjunctive Bayesian Networks (CBNs) \cite{gerstung2009quantifying,sakoparnig2012efficient,hosseini2019estimating} and Suppes-Bayes Causal Networks (SBCNs) \cite{ramazzotti2015capri,de2016tronco,ramazzotti2018modeling}. 
Such models capture the temporal ordering and the conditional dependencies among the events, and can be inferred by pooling data of multiple patients. 

For instance, cross-sectional mutational profiles of cancer patients generated, e.g. via variant calling from bulk sequencing data, can be used to infer the most likely trends of accumulation of somatic mutations during the development of a certain cancer type. These trends enable the \emph{prediction} of the the next step(s) of the disease evolution of any given patient, with evident implications on prognostic strategies and therapeutic interventions \cite{wang2015predictive}. 
Yet, the inference problem is complicated by the high levels of heterogeneity typically observed in most tumor types, which are due to the existence of multiple independent evolutionary trajectories, often involving shared subsets of events with complex dependencies  \cite{turajlic2019resolving}.  

In this regard, most existing computational approaches allow one to model conjunctive processes (i.e. \texttt{AND} logical operator), according to which a certain event can occur only if all its direct ancestors (i.e. parent events) have occurred. 
This strategy allows one to cover a wide range of evolutionary processes observed in real-world scenarios, yet it struggles when more complex dependencies are present, such as \emph{disjuctions} (i.e. \texttt{OR}), according to which a certain event can occur if at least one of its parents has occurred, or \emph{mutual exclusivity} relations (i.e. \texttt{XOR}), in which a given event can occur if only one of its parents has occurred, e.g. due to synthetic lethality \cite{o2017synthetic}. 

Therefore, some methods have been recently devised to overcome the limitations of conjunctive models, by allowing one to assess the existence of arbitrary logical formulas connecting events \cite{ramazzotti2015capri,ramazzotti2018modeling}.
However, such approaches require that such formulas are identified prior to the inference and provided as input, thus requiring either to possess a deep biological knowledge on the underlying evolutionary process, or to employ ad hoc computational strategies to identify specific patterns among events \cite{ciriello2012mutual}. 

In this work we propose \alg{} (Predictive Models of Cancer Evolution), a computational framework for the inference of expressive probabilistic graphical models of cancer evolution from cross-sectional mutational profiles of cancer samples, which are named Hidden Extended Suppes-Bayes Causal Networks (HESBCNs).  
The main novelty of \alg{} is the automated identification of logical formulas connecting the events, which is achieved via an efficient  Markov Chain Monte Carlo (MCMC) search scheme. 
In addition, \alg{} employs an Expectation-Maximization (EM) strategy on a continuous-time Hidden Markov Model (HMM), in order to assign an error probability to the dataset and an evolutionary time (i.e. the expected waiting time required to observe a given sample) to any node of the output model. 
On the one hand, this allows one to evaluate the \emph{predictability} of any given tumor, as initially  proposed in \cite{hosseini2019estimating}, and which roughly measures the repeatability of the  evolutionary patterns observed in a given tumor  \textcolor{black}{by quantifying the  entropy of the different trajectories of the model}. 
On the other hand, by attaching samples to the evolutionary paths of the output model, it is possible to estimate the evolutionary time of any given sample with respect to the expected progression of the corresponding tumor type (notice that time is measured in arbitrary units and the temporal position of the samples is relative to that of the other samples of the dataset). 

We assessed the performance of our approach via extensive simulations, in which we tested the capability of \alg{}, HCBNs \cite{hosseini2019estimating} and standard Bayesian Networks \cite{koller2009probabilistic} in recovering the ground-truth topology, with respect to generative models with distinct sample size, topological complexity and noise. \alg{} significantly outperforms competing methods in most settings, proving its superior expressivity and robustness to noise. 

Finally, we applied \alg{} to $7866$ samples from $16$ cancer types from The Cancer Genome Atlas (TCGA) database \cite{weinstein2013cancer}. 
In addition to the evolutionary models, we computed the predictability of each cancer type, also assessing the possible correlation with the overall mutational burden. 
Importantly, we executed a combined regularized Cox regression  \cite{simon2011regularization,tibshirani2012strong} and Kaplan-Meier survival analyses by employing the patient-specific evolutionary paths and evolutionary times inferred from the HESBCN models. This allowed us to identify $7$ cancer types in which a risk-based stratification of patients defines statistically significant differences in the overall survival. 
This important result proves that \alg{} can be employed to generate experimental hypotheses with translational relevance and high prognostic power and which might, in turn, drive the design of cancer- and patient-specific therapeutic strategies.

\section{Methods}


    
    

Bayesian Networks (BNs) \cite{koller2009probabilistic} have been often employed to model the likely temporal trends of accumulation of somatic variants in cancer, in many cases by fitting binarized mutational profiles generated from cross-sectional bulk sequencing experiments \cite{beerenwinkel2015cancer,schwartz2017evolution,10.1371/journal.pcbi.1007246}.
The different existing approaches are characterized by distinct levels of expressivity, ranging from tree models \cite{desper1999inferring,loohuis2014inferring}, to conjunctive Bayesian networks (CBNs) \cite{beerenwinkel2007conjunctive,gerstung2009quantifying} and to more complex representations involving logical formulas among genomic events, including Suppes-Bayes Causal Networks (SBCNs) \cite{ramazzotti2015capri,caravagna2016algorithmic,bonchi2017exposing,gao2018causal,bonchi2018probabilistic} and Extended Suppes-Bayes Causal Networks (ESBCNs) \cite{ramazzotti2018modeling,ramazzotti2019efficient}.  
In particular, since the search of logical formulas implies an exponential growth in computational complexity, all available algorithms require a set of logical formulas as input, which are then tested in the search step \cite{ramazzotti2018modeling,ramazzotti2019efficient}.

In this work, we first introduce the Hidden Extended Suppes-Bayes Causal Network (HESBCN) model, in which a HMM is added to ESBCNs to simulate the stochastic processes related to the accumulation of genetic alterations during cancer development, as well as the measurement error. In particular, we assume that the time between two (independent) events is exponentially distributed with rate $\lambda_i$, whereas measurement errors are modeled via a Bernoulli process with error probability $\epsilon$. 

\begin{figure}
\center
  \includegraphics[width=0.80\textwidth]{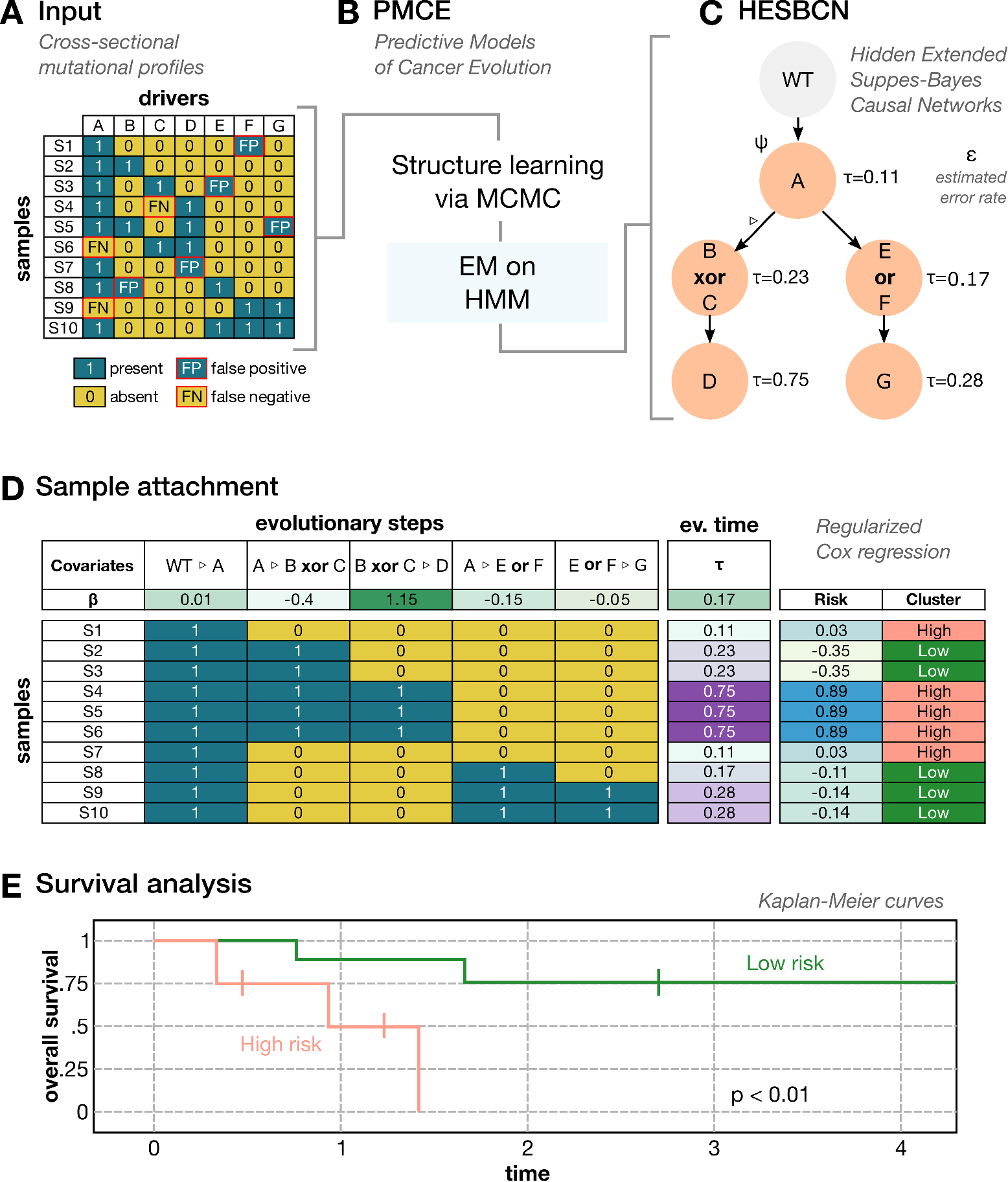}
  \caption{{\bf \alg{} framework}. {\bf (A)} \alg{} takes as as input binarized cross-sectional profiles of driver mutations, generated via bulk sequencing experiments. Data can include false positives and false negatives. {\bf (B)} \alg{} employs a Monte-Carlo Markov Chain (MCMC) search and an Expectation-Maximization step to infer the Maximum a Posteriori (MAP) structure and the parameters of {\bf (C)} a Hidden Extended Suppes-Bayes Causal Network (HESBCN), which describes the evolutionary steps $\triangleright$ from the wild-type (WT). \alg{} is capable of inferring logical formuals $\psi$ connecting the events, the evolutionary time $\tau$ of each event, and the error rate $\epsilon$ of the dataset. {\bf (D)} \alg{} attaches the samples to the HESBCN model via maximum likelihood estimation, and employs the evolutionary steps and time in order to perform a Regularized Cox Regression  \cite{simon2011regularization,tibshirani2012strong} on survival data. This allows one to identify the relevant covariates and the corresponding risk coefficients $\beta$. Samples are then stratified in clusters according to the overall risk. {\bf (E)} Risk clusters are compared via Kaplan-Meier survival analysis.   }
\label{fig:esempio}
\end{figure}

We then present a new algorithmic framework named \alg{} for the inference of HESBCNs from cross-sectional mutational profiles of cancer samples, which includes a two-step procedure: $(i)$ a MCMC search to infer the Maximum a Posteriori (MAP) structure of HESBCNs and the concomitant automated inference of logical formulas, $(ii)$ an EM procedure to infer the parameters of the HMM (see Figure \ref{fig:esempio}).





More in detail, a HESBCN is a \emph{probabilistic graphical model} \cite{pearl2009causality} defined by 
\begin{itemize}
    \item the set $\{N\}=\{\psi_1,\dots,\psi_n\}$ of vertices representing logical formulas involving one or more atomic event(s) of a model. In our case, the atomic events are binary variables modeling the presence/absence of genomic variants (e.g. single-nucleotide or structural variants), whereas logical relations are limited to \texttt{AND}, \texttt{OR} (soft exclusivity) and \texttt{XOR} (hard exclusivity). 
    
    \item  The set $\{\triangleright\} = \{\triangleright_1, \dots, \triangleright_w\}$ of $w$ edges representing the conditional dependencies among the vertices of the model (e.g. $\psi_i\triangleright \psi_j$) and which are also referred to as \emph{evolutionary steps} in the case studies. 
\noindent
We note that, by construction, HESBCNs model Direct Acyclic Graphs (DAGs), in which a vertex can have more than one parent and models with multiple roots and/or disconnected components are allowed.
 
    \item The set of conditional probabilities associated to the vertices, $\{\theta\}=\{\theta_1,\dots, \theta_n\}$, where  $\theta_i$ is the conditional probability that a logical formula $\psi_i\in N$ is true, given that its predecessor formulas are true. 
   
    \item The error probability $\epsilon$ of the dataset, by assuming that the false positives/negatives rates in the input binary data are modeled with a Bernoulli  process, as in \cite{sakoparnig2012efficient}. In our case, we assume that the rate of false positives and false negatives is identical and $=\epsilon/2$.   
   
    \item The set $\{\lambda\}=\{\lambda_1,\dots,\lambda_n\}$ of rates of the Poisson processes of the continuous-time HMM, associated to the vertices of the model, which allow one to estimate the expected waiting time of a node, given that its predecessor has occurred. 
\end{itemize}

\noindent
\subsection{Structure learning of HESBCNs via MCMC}
The \alg{} framework first aims at inferring the Maximum a Posteriori (MAP) HESBCN structure.   
Given a cross-sectional mutational profiles dataset $D$, generated, e.g. from bulk sequencing data, let $r$ be the total number of atomic events, i.e. the genomic variants observed in at least one sample. 
Let be $d^i \in\{0,1\}^r$ the binarized mutational profile of length $r$ of the $i^\text{th}$ sample belonging to the input dataset $D$, such that  $d^i_j=1$  if the $j^\text{th}$ mutation is detected in sample $i^\text{th}$, $0$ otherwise (a discussion on the binarization of variant allele frequency profiles from sequencing data can be found in \cite{ramazzotti2015capri,caravagna2018detecting}).

Assuming that all the observations are independent, the posterior probability of a HESBCN model ($\{\triangleright\}$, $\{\theta\}$, $\epsilon$), given the data $D$, is proportional to:

  \begin{equation}
  \begin{split}
     &           P(\{\triangleright\},\{\theta\},\epsilon|D) \propto \\
    & \propto\left[ \prod_{d^i\in D}\sum_{\theta_k\in\{ \theta \}}P(d^i|\{\triangleright\},\theta_k,\epsilon)\right]\cdot P(\{\triangleright\}) P(\{\theta\})  P(\epsilon),\\
    \end{split}
    \label{eq:posterior_MCMC}
   \end{equation}
\noindent
where: $P(\{\theta\}) = \prod_{i=1}^n{P(\theta_i)}$ and $P(\theta_i)$ are Beta-priors with both shape parameters equal to $10^{-5}$, as in \cite{sakoparnig2012efficient};
  $P(\epsilon)$ is the prior for the error probability, chosen as uniform in the structure learning step;
 $P(\{\triangleright\})$ is the prior for the network structure, which here encodes the conditions of Suppes' probabilistic causation \cite{suppes1973probabilistic}, as follows: 

\begin{equation}
   P(\{\triangleright\})=\prod_{i\in \{N\}}
\begin{cases}
1 & \quad \text{if}\quad \forall \psi_j \in \{Pa(\psi_i)\} \\
\quad & P(\psi_j)>P(\psi_i) \quad
\text{and} \\
\quad & P(\psi_i|\psi_j)>P(\psi_i|\neg{\psi_j})\\
 0 & \quad \text{otherwise}.\\
\end{cases} 
\end{equation}
where $\{Pa(\psi_i)\}$ indicates the parent set of the formula $\psi_i$.
Notice that the convergence to the MAP structure in this scenario was demonstrated in \cite{ramazzotti2015capri}. 

 The likelihood term $P(d^i|\{\triangleright\},\theta_k,\epsilon)$ represents the  probability to observe the  $d^i$ sample given a HESBCN model. This term is evaluated for each node $k=1,\dots,n$  by first solving  each formula  from the data ($=1$ if the formula is true, $=0$ otherwise) and then computing the conditional probability $\theta_k$.


   In order to infer the structure of a HESBCN given the data, \alg{} employs a Monte Carlo Markov Chain (MCMC) approach with a Gibbs' sampling scheme from the posterior distribution, which is a modification of that proposed in \cite{sakoparnig2012efficient}. The MCMC starts from a randomly initialized HESBCN structure and includes $8$ different moves:\\
   \begin{enumerate}
       \item {\it Modify Theta}: given a randomly chosen node $i$, $\theta_i$ is set to a new value, which is sampled from a uniform distribution: $[0,1]$.
       \item {\it Create a new parental relation}: a new parental relation is randomly added to the model, by avoiding the creation of cycles.
       \item {\it Delete a parental relation}: a  randomly chosen parental relation is deleted from the model. 
       \item {\it Add an edge into  a transitive closure relation}: we select a node, and considering its transitive closure a new  edge is inserted into the model.
       \item {\it Delete an edge from  a transitive closure relation}: an  edge is deleted from the model, by considering only the edges belonging to the transitive closure of a given node.
       \item {\it Node swap}: two randomly chosen nodes are swapped.  
       \item {\it Reincarnation}: a {\it delete a parental relation} move is
       followed by a {\it create a new parental relation} move. 
       \item {\it Local restart}: the inference is restarted from a new random configuration, if the previous MCMC has converged to a deadlock.
    \item [*] During the MCMC, every time that a node (e.g. \textit{A}) happens to have more than one parent (e.g. \textit{B} and \textit{C}) a logical operator is chosen with uniform probability among \texttt{AND}, \texttt{OR} and \texttt{XOR} and associated to the elements of its parent set (e.g. $(\textit{B} \quad \texttt{XOR} \quad \textit{C}) \triangleright \textit{A}$).

   \end{enumerate}
   
The acceptance probability $\rho$ of a proposed MCMC sample, is given by:

      \begin{equation}
     {\scriptstyle
       \rho=\min \Bigg\{ 1, \frac{  P(\{\triangleright\},\{\theta\},\epsilon|D)'\times \text{MSP}(\{\triangleright\},\{\theta\},\epsilon)'\times \text{TP}(\{\triangleright\},\{\theta\},\epsilon|\{\triangleright\}',\{\theta\}',\epsilon) }{  P(\{\triangleright\},\{\theta\},\epsilon|D)\times \text{MSP}(\{\triangleright\},\{\theta\},\epsilon)\times \text{TP}(\{\triangleright\}',\{\theta\}',\epsilon|\{\triangleright\},\{\theta\},\epsilon)   }\Bigg\}}
   \end{equation}
where $'$ indicates the quantities associated with the structure after the proposed move. MSP stands for move selection probability (see Table \ref{tab:mps} for default probabilities). TP is the transition probability from the current structure to the new one, given the selected move. 
It is important to notice that despite being in the equation, the error rate $\epsilon$ is set to $0$ during the structure inference.

In order to limit the size of the logical formulas included in the output model, \alg{} maximizes a score composed by Equation \eqref{eq:posterior_MCMC} plus a regularization term, e.g. BIC or AIC.
As output, \alg{} returns the MAP HESBCN structure and parameters retrieved after a user-selected number of MCMC iterations and restarts. 

\begin{table}
\caption{Values of move selection probability (MSP) for the moves described in the Methods section.}
\centering
\label{tab:moves}
\begin{tabular}{l  l}
Move & MSP default value \\
\hline
{\it  Modify theta} & $ 0.15$ \\
{\it Create a new parental relation} & $0.4$ \\
{\it Delete a parental relation}& $0.2$\\
{\it Add an edge into  a transitive closure relation} & $0.04$ \\
{\it Delete an edge from  a transitive closure relation} & $0.04$  \\
{\it Node swap} & $ 0.07$ \\
{\it Reincarnation} & $ 0.1$ \\

\hline
\end{tabular}
\label{tab:mps}

\end{table}

 \subsection{Hidden Markov Model}
 \label{sec:lambda}
  The accumulation of genomic variants during cancer evolution can be modeled as a stochastic time-dependent process, e.g. via a continuous-time Hidden Markov model, as originally proposed in \cite{gerstung2009quantifying}.
 More in detail, let us suppose that the formula $\psi_i$ of a HESBCN model is satisfied after time $t_i$. A model composed by $n$ nodes will include $n$ waiting times $t_1,\dots, t_n$.
As commonly done with stochastic branching processes, we assume an exponentially distributed time and, accordingly, we define the waiting time associated to any logical formula $\psi_i$ as: 
\begin{equation}
    t_i=\max\limits_{\psi_j \in \{Pa(\psi_i)\}}{t_j}+Z_i,\qquad Z_i\sim exp[\lambda_i],
\end{equation}
where $\lambda_i$ is the rate of the exponential distribution of the $i$-th formula. 

Let us then define a {\it logical formula path} as an ordered sequence of logical formulas:
\begin{equation}
    \mathbf{\Psi}_l=(\dots(\psi_1\triangleright\psi_2)\triangleright\dots)\triangleright\psi_l,
    \label{eq:DNF-path}
\end{equation}
such that $\psi_{i-1} \in \{Pa(\psi_i)\},\quad \forall i=2,\dots,l$ .
If we suppose that the probability density of a logical formula path $\mathbf{\Psi}_l$ factorizes according to $P(\mathbf{\Psi}_l)=\prod_{i=1}^l P(\psi_i)$, then the following equation holds:
\begin{equation}
    P(\mathbf{\Psi}_l)=\prod_{i=1}^l\int_0^{\infty}dt_i\int_0^\infty dt_s \chi(t_i,t_s)f(t_i|\{t_j\}_{\{Pa(\psi_i)\}})f(t_s),
    \label{eq:posterior}
\end{equation}
where $t_s$ is time of the diagnosis, $\chi(t_i,t_s)=1$ if $t_s> t_i$, $0$ otherwise, $f(t_s)$ is the probability density of the diagnosis time (which is assumed to be uniform) and $f(t_i|\{t_j\}_{\{Pa(\psi_i)\}})$ is the probability density of $t_i$, conditioned on its predecessors.

In principle, one could directly maximize Eq. \eqref{eq:posterior} to estimate the set of optimal $\{\lambda_i\}$. 
However, dealing with real data, it may be sound to account for  experimental noise, i.e. the false positives and false negatives possibly included in the input data $D$.  

Thus, let us introduce the set of theoretical genotypes $\{\mathcal{G}_{\mathbf{\Psi}_l}\} = \{g^1, \dots, g^q$\} as the set of $q$ possible genotypes subsumed by the HESBCN path ${\Psi}_l$ and that are represented as ternary vectors: $g^k=\{0,1,*\}^r$, such that $g^k_j=1$ if the variant $j$ is present in the theoretical genotype $k$, $g^k_j=0$ if it is absent, $g^k_j=*$ if the variant is not included in path ${\Psi}_l$ and, therefore, it can either be present or absent.

Then, we model the measurement errors via a  Bernoulli process with an error probability $\epsilon$. In detail, the probability of observing the genotype $d^i(r)$, given that the theoretical genotype is $g^k(r)$ and the error probability is $\epsilon$, is given by:
    \begin{equation}
        P(d^i(r)|g^k(r),\epsilon)=\epsilon^{\text{HD}(g^k(r),d^i(r))}(1-\epsilon)^{r-\text{HD}(g^k(r),d^i(r))}.
        \label{eq:hamming}
    \end{equation}
  Here, $ \text{HD}( g^k(r),d^i(r)  )$ denotes the Hamming distance between
   $d^i(r)$ and  $g^k(r)$.\\

Accordingly, the posterior probability of a HESBCN model becomes:
\begin{equation}
\begin{split}
    &P(\{\triangleright\},\{\theta\},\{\lambda\},\epsilon|D) \equiv
 P(\{\mathcal{P}\},\{\lambda\},\epsilon|D)= \\
    &\sum_{{\mathbf{\Psi}_l} \in \{\mathcal{P}\}}
   \prod_{d^i(r)\in D} \frac{  P(\mathbf{\Psi}_l)  P(d^i(r)|\{\mathcal{G}_{\mathbf{\Psi}_l}\},\epsilon)}{  \sum_{\mathbf{\Psi}_l\in \{\mathcal{P}\}} P(\mathbf{\Psi}_l)  P(d^i(r)|\{\mathcal{G}_{\mathbf{\Psi}_l}\},\epsilon)},
    \label{eq:DNF}
\end{split}
\end{equation}
 where  $\{\mathcal{P}\}$ is set of the  logical formula paths included in the HESBCN model, $P(\mathbf{\Psi}_l)$ is the probability density from Eq. \eqref{eq:posterior} and the likelihood is defined by assuming that the probability of a logical formula path is equal to the product of the probability of every related theoretical genotypes, as follows: 
\begin{equation}
    P(d^i(r)|\{\mathcal{G}_{\mathbf{\Psi}_l}\},\epsilon)=\prod_{g^k(r)\in\{\mathcal{G}_{\mathbf{\Psi}_l}\}} P(d^i(r)|g^k(r),\epsilon),
\end{equation}

To estimate the sets of $\{\lambda_i\}$ and $\epsilon$ while the structure is kept fixed, we employ a EM algorithm to maximize equation \eqref{eq:DNF}.
Thanks to this procedure, \alg{} returns the expectation time $\lambda_i$ for each node of the model (notice that the choice of not including the search of $\lambda_i$ and $\epsilon$ directly in the MCMC was made to reduce the overall computational complexity and speed the computation up). 

\alg{} then computes the maximum likelihood (ML) attachment of the samples to the HESBCN model using Eq. \eqref{eq:hamming} with the $\epsilon$ estimated in the EM step. This allows one to return: $(i)$ the ML theoretical genotype of each sample and $(ii)$ the ML evolutionary steps, i.e. the set of parental relations among true logical formulas of that sample (see Figure \ref{fig:esempio}). 

Finally, since stochastic branching processes are additive, it is possible to compute the evolutionary time of each sample as $\tau = \sum_{s}\frac{1}{\lambda_s}$, i.e. the expected waiting time to cover all the related evolutionary steps (measured in arbitrary time units, which can be rescaled with respect to the diagnosis time $t_s$, as in \cite{sakoparnig2012efficient}).

\subsection{Predictability}
After inferring the HESBCN model from the mutational profiles of the samples of a given tumor, it is possible to quantify its {\it predictability} \cite{estrada2007statistical,hosseini2019estimating}, \textcolor{black}{that is a quantity directly related to the entropy computed considering the probability of all the possible evolutionary paths of the model.}



More in detail, if each node of a HESBCN is a state of the evolutionary history of a given tumor type, then it is natural to interpret the set of the  logical formula paths $\{\mathcal{P}\}$  as the set of all the possible evolutionary trajectories.
Since $1/\lambda_i$ is the expected waiting time of the $i$-th node, it is intuitive to relate the probability of a single logical formula path of length $l$ to the set of $\lambda$ associated to the nodes belonging to the path, i.e. \cite{estrada2007statistical,hosseini2019estimating}:
\begin{equation}
    \Pi(\mathbf{\Psi}_l)=\prod_{i=1}^l\frac{\lambda_{\psi_i}}{\sum_{\psi_h|\psi_i\in \{Pa(\psi_h)\}}\lambda_{\psi_h}}.
\end{equation}
%
It is then possible to define the entropy of the set of logical formula paths of a HESBCN model $\{\mathcal{P}\}$ as \cite{estrada2007statistical,szendro2013predictability}:
\begin{equation}
    H_{\{\mathcal{P}\}}=-\sum_{\mathbf{\Psi}_l\in \{\mathcal{P}\}}\Pi(\mathbf{\Psi}_l) \log (\Pi(\mathbf{\Psi}_l)).
    \label{eq:entropy}
\end{equation}

This quantity estimates the amount of uncertainty in selecting a certain path among all possible paths of a given model.
As a consequence, a HESBCN model has a maximum entropy if the probability associated to any possible path of the process is the same:
 $H_{max}=\log(l!)$ \cite{hosseini2019estimating}.\\ 
Intuitively, entropy in Eq.\eqref{eq:entropy}  could be  used to quantify how much of the evolutionary process is “localized” in a few paths.
The predictability of a HESBCN model characterized by a set of logical formula paths $\{\mathcal{P}\}$, is then defined as in \cite{hosseini2019estimating}:
\begin{equation}
    \Phi_{\{\mathcal{P}\}}=1-\frac{H_{\{\mathcal{P}\}}}{H_{max}}.
    \label{eq:pred}
\end{equation}

As a matter of fact, $0\leq\Phi_{\{\mathcal{P}\}}\leq 1$. If $\Phi_{\{\mathcal{P}\}}=0$, all possible logical formula paths are equally probable, whereas, if $\Phi_{\{\mathcal{P}\}}=1$, only one path is possible, which indicates perfect predictability.
\noindent

\subsection{Survival Analysis}
The evolutionary path and time of a given sample can be employed as risk estimators, in order to stratify patients into risk groups.
To assess the prognostic power of the cancer evolution models returned by \alg{}, we implemented a combined regularized Cox regression  \cite{simon2011regularization,tibshirani2012strong} and Kaplan-Meier survival analysis.

In brief, according to the standard Cox proportional hazards model \cite{cox1972regression}, given a baseline of risk $r_0$ of a given disease, and a vector of covariates $X_i$, the risk associated to $X_i$ is given by the hazard function:
\begin{equation}
    r(t)=r_0(t)e^{\sum_i \beta_i X_i },
\end{equation}
where $\beta_i$ are coefficients that measure the impact of every covariate (if $\beta_i>0$ the covariate is associated to a risk increment).
In our case, we applied the regularized Cox regression via Coxnet  \cite{simon2011regularization,tibshirani2012strong}, which allows one to select the subset of covariates that minimize the cross validation error, by employing the elastic net regularizer (see Supplementary Figure S1). 

In particular, we employed as input covariates for the analysis: $(i)$ all the parental relations between the nodes (i.e. the formulas $\psi_i$) of the MAP HESBCN model, assessed with respect of the maximum likelihood evolutionary steps associated to each sample ($=1$ if both the formulas in that relation are satisfied, $=0$ otherwise), and $(ii)$ the evolutionary time $\tau$ of each sample (in arbitrary units, see Figure \ref{fig:esempio}). 

For each tumor, we selected the set of $\beta_i$ associated to the minimum cross-validation error and kept as significant only the cancer types displaying at least one covariate with non-zero $\beta_i$. 
We then computed for each sample a risk score as follows: 
\begin{equation}
 \xi = \sum_i \beta_i X_i. 
\label{eq:risk}
 \end{equation}
 
This allowed us to stratify the patients into three different risk clusters: $(i)$ high risk with $\xi>0$, $(ii)$ medium risk with $\xi=0$ (baseline, typically including all the samples with impactless covariates), $(iii)$ low risk with $\xi<0$. 
The survival curves of the outcoming risk clusters were finally assessed via standard Kaplan-Meier analysis.

\section{Results}
\subsection{Results on simulations}

\begin{figure}
\centering
  \includegraphics[width=1.00\textwidth]{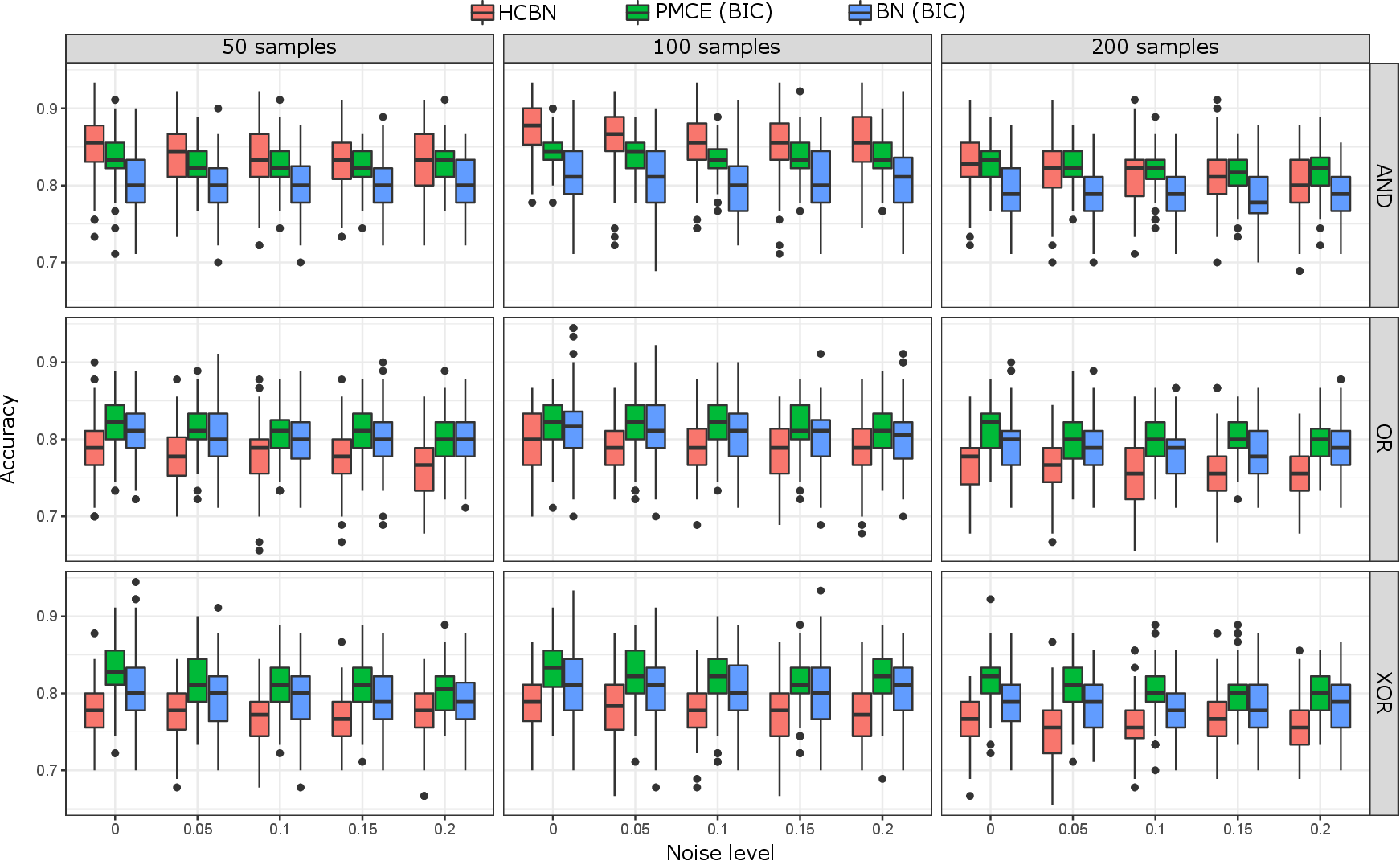}
\caption[.]{ {\bf Performance assessment on simulations.} $4500$ independent synthetic datasets were generated from DAG generative topologies with $m = 10$ nodes (representing e.g. genomic variants) and density $=0.4$, three distinct levels of topological complexity (only \texttt{AND} -- panels {\bf (A)},  \texttt{OR} -- panels {\bf (B)} and \texttt{XOR} -- panels {\bf (C)}, respectively), three sample sizes ($m = [50, 100, 200]$) and $5$ noise levels ($\epsilon = [0, 0.05, 0.10, 0.15, 0.20]$). \alg{} was compared with HCBNs \cite{gerstung2009quantifying} and standard BNs \cite{koller2009probabilistic} (with BIC regularization) on overall accuracy with respect to the ground-truth generative topology. Box-plots return the distributions on all simulations. 

}
\label{fig:synth_res}
\end{figure}
 To assess the performance of \alg{} with respect to competing approaches, we performed extensive tests on simulations. 
 Synthetic datasets were sampled from randomly generated weekly connected DAGs with $r = 10$ atomic events and density $=0.4$. 
 Three levels of topological complexity were implemented and, in particular, $100$ DAGs were generated to include only \texttt{AND} logical relations (associated to the confluences), $100$ DAGs only \texttt{OR}'s and $100$ DAGs only \texttt{XOR}'s, for a total of $300$ generative topologies.   
 For each topology, $3$ different values of sample size ($m = [50, 100, 200]$) and $5$ levels of noise rates ($\epsilon = [0,0.05, 0.10, 0.15, 0.20]$) were scanned to generate a synthetic mutational profile dataset, for a total of $4500$ independent datasets. 
 We compared \alg{} with Hidden Conjunctive Bayesian Networks (HCBNs) \cite{gerstung2009quantifying} and with standard Bayesian Networks (BNs) \cite{koller2009probabilistic} and compared the performance of the methods by assessing the accuracy $\frac{\text{TP} + \text{TN}}{\text{TP} + \text{FP} + \text{TN} + \text{FN}}$ with respect to the ground-truth generative topology. 
 Both \alg{} and standard BNs were executed with BIC regularization. 

In Figure \ref{fig:synth_res}, one can see that \alg{} significantly outperforms the competing approaches in all settings, except for the scenarios with high samples size and \texttt{AND} generative topologies, in which, as expected, HCBNs display a superior accuracy, likely due to the algorithmic design aimed at specifically inferring conjunctive relations and the smaller search space.

The most difficult setting appear to be that involving \texttt{XOR} generative topologies (Figure \ref{fig:synth_res}{\bf C}), in which consistently lower performances are observed for all methods, with respect to synthetic datasets generated with distinct topological complexity. This result is likely related to the presence of spurious dependencies among events due to the properties of the \texttt{XOR} logical formula. 

Importantly, the overall performance of \alg{} is only slightly affected by even significantly high levels of false positives/negatives in the data (in all setting and with all topologies), and this is especially true for configurations with large sample size. This result proves the robustness of our approach in delivering reliable predictions from real-world data. 

\subsection{Results on real data}
We applied \alg{} to $7866$ samples from $16$ cancer types from The Cancer Genome Atlas (TCGA) database \cite{weinstein2013cancer}. The dataset includes bulk sequencing data, which is often coupled with clinical information on the related patient. In particular, the HESBCN model was inferred separately for every cancer type via \alg{}, by considering putative driver mutations only, as identified in related works \cite{bailey2018comprehensive}. 
All models are displayed in Supplementary Figures S2-S17 and highlight different degrees of complexity and heterogeneity.  

\begin{figure*}
\centering
      \includegraphics[width=1.00\textwidth]{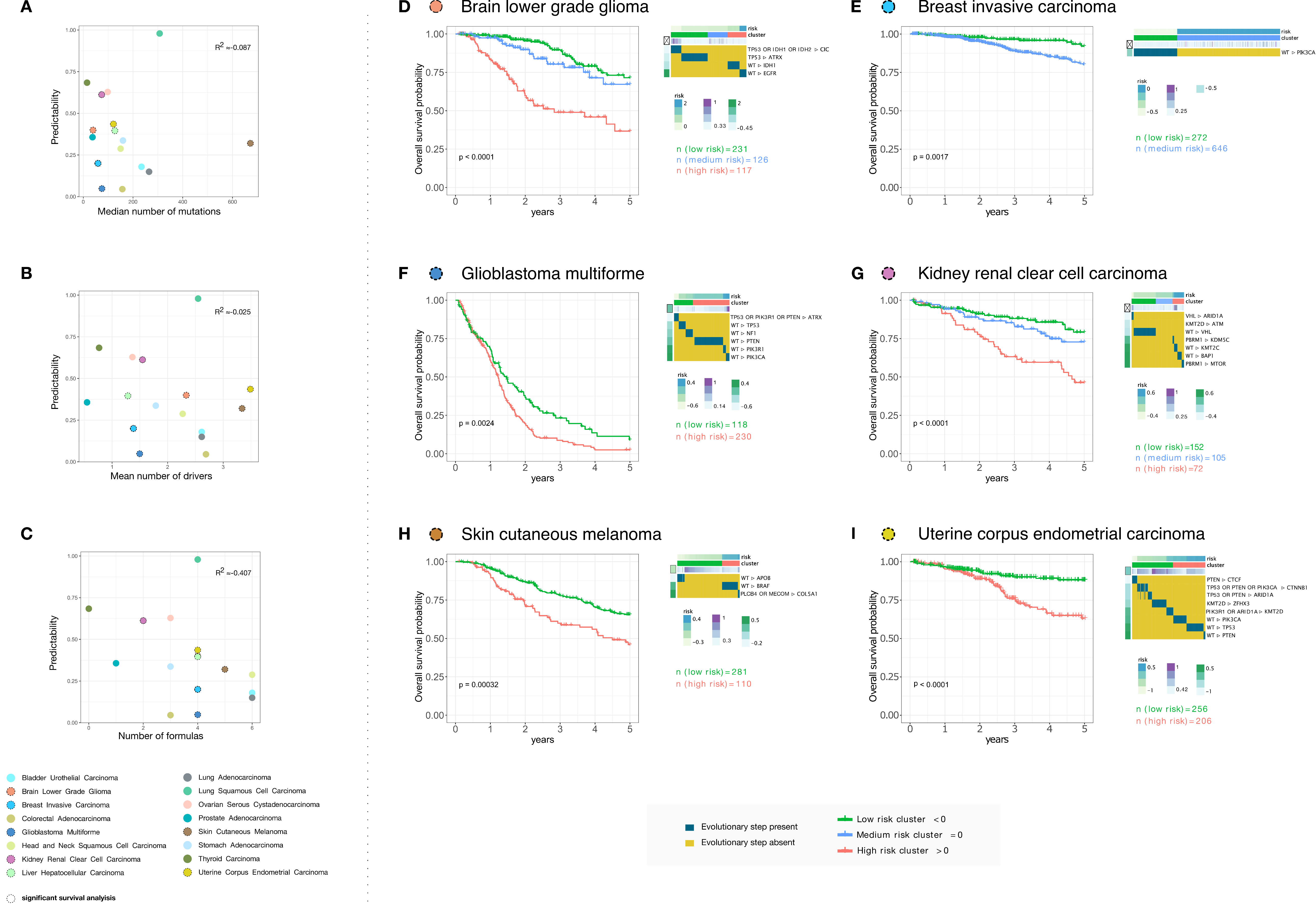}
\caption{{\bf Application of \alg{} to 7866 TCGA samples.} \alg{} was applied to $7866$ samples of $16$ different cancer types from the TCGA database \cite{weinstein2013cancer}, on the binarized profiles of driver mutations from \cite{bailey2018comprehensive}. {\bf (A)} Scatter-plot returning the relation between the predictability of each tumor type and the median number of mutations, {\bf (B)} the mean number of drivers, \textcolor{black}{{\bf (C)} the number of logical formulas present in the HESBCN model}. The $7$ cancer types showing a significant survival analysis p-value (see Methods) are circled with a dashed line. The stratification of patients in risk groups based on evolutionary steps and time obtained via \alg{} and selected via Regularized Cox Regression is shown with Kaplan-Meier survival plots for {\bf (D)} brain lower grade glioma, {\bf (E)} breast invasive carcinoma, {\bf (F)} glioblastoma multiforme, {\bf (G)} kidney renal clear cell carcinoma, {\bf (H)}  skin cutaneuos melanoma {\bf (I)} uterine corpus endometrial carcinoma. For all tumor types, the selected covariates, the sample size of the clusters, and the Kaplan-Meier p-value are shown. In addition, a heatmap returns the presence/absence of the selected covariates, the value of $\beta_i$ for each covariate and the values of $\tau$ of each sample.}
\label{fig:pred}
\end{figure*}

\paragraph{\textcolor{black}{Predictability.}} For each model, we evaluated the predictability as per Eq. \eqref{eq:pred} and the results are shown in Figure \ref{fig:pred}{\bf A-C}. One can notice that $\Phi$ is significantly different across cancer types. For instance, lung squamous cell carcinoma display a value of $\Phi\approx 0.97 $, which is likely due to the fact that most patients follow a single path including a driver mutation on \textit{TP53}.  Conversely, both glioblastoma multiforme and colorectal adenocarcinoma show a very low score of $\Phi \approx 0.05 $, hinting at the presence of multiple independent evolutionary trajectories in distinct patients. 

We investigated the possible correlation between the predictability and the overall mutational burden, measured via both the median number of total mutations and the mean number of driver mutations (see Figure \ref{fig:pred}{\bf A-B}). Notably, the overall correlation appear to be limited in both cases: $R^2=-0.087$ and $R^2=-0.025$, respectively.
Despite a higher number of driver mutations may imply, in principle, a larger number of possible evolutionary trajectories, the overall tumor burden appears not to be a clear indicator of predictability, pointing at the existence of preferred routes of cancer evolution for certain cancer types, as opposed to the limited regularities that characterize the evolution of distinct cancer types \cite{burrell2013causes,mcgranahan2015biological,turajlic2018tracking}.

\textcolor{black}{Conversely, the number of formulas included in the evolution models shows a moderate anti-correlation with the predictability ($R^2=-0.407$, Figure \ref{fig:pred}{\bf C}). This result would suggest that, as intuitively expected, tumors with a larger number of possible trajectories are indeed the less predictable.}

\textcolor{black}{We also assessed the contribution of each genomic alteration included in the models to the overall predictability of the respective tumors. 
In particular, for each tumor type and each genomic event, we computed the predictability value of the subgraph defined by considering only the paths from the root to that event.}
\textcolor{black}{With this analysis we obtained a predictability score for each genomic alteration with respect to every cancer type, which is displayed as a heatmap in Supplementary Figure S20.}

\textcolor{black}{
Such analysis highlighted that driver mutations involved in many evolutionary paths, i.e. representing an early or necessary mutational event for a given tumour type, typically provide limited information in terms of predictability for that cancer. The most evident example is represented by the mutation of \textit{TP53}, that is a pivotal driver in most cancer types, which shows always a very limited predictability score (see Supplementary Figure S20). The same pattern was observed for tissue-specific drivers, see, for instance, the mutations hitting \textit{APC} in colorectal cancers, \textit{BRAF} in thyroid cancers, \textit{IDH1/2} in brain lower grade gliomas or \textit{VHL} and \textit{PBRM1} in kidney carcinomas. Interestingly, we also found driver mutations showing a mixed behaviour, being pivotal and scarcely contributing to predictability in some cancer types, but defining high predictable subtypes in others. A very interesting example in this regard is the mutation of \textit{PIK3CA}, that is a major cancer driver of ER+ breast carcinomas, which displays a very low predictability in breast cancer, but a very high predictability score, e.g. in brain lower grade gliomas and stomach cancers.}

\paragraph{\textcolor{black}{Survival analysis.}}
Survival analysis was executed by employing the Regularized Cox Regression via Coxnet \cite{simon2011regularization, tibshirani2012strong} with $10000$ maximum iterations via elastic net, which allowed us to identify the significant subset of covariates for each tumor type.
In particular, for every sample, we employed as covariates: $(i)$ each evolutionary step (i.e. parental relation of the HESBCN model) present in the maximum likelihood theoretical genotype and $(ii)$ the evolutionary time $\tau$.
In detail, for $7$ cancer types we found an non-empty set of covariates (i.e. $\beta \neq 0$) associated to the minimum cross-validation error (See Supplementary Figure S1 and Supplementary File 1).
The samples of each of such cancer types were then divided into three risk clusters, according to Eq. \eqref{eq:risk}. 

\textcolor{black}{We first estimated the stability of such results, by performing cross-validation as follows: $(i)$ we randomly selected 80$\%$ of the samples of each cancer type, $(ii)$ we performed the regularized Cox regression analysis for each sampled dataset and estimated the $\beta$ coefficients, $(iii)$ we independently repeated the sampling procedure $1000$ times;  $(iv)$ we finally computed the Pearson correlation coefficient among the $\beta$ coefficients obtained from the sampled dataset and those from the full dataset, for each of the $1000$ independent samplings.}
\textcolor{black}{As a result, the median Pearson correlation coefficients for the 7 considered cancer types are all larger than $0.90$ (i.e. high correlation), proving the significant stability of the analysis (full results are provided in Supplementary Figure S19).}

In Figure \ref{fig:pred}{\bf D-I}, we show the Kaplan-Meier plots for $6$ cancer types, in addition to the selected covariates and the associated risk coefficient (since in the liver hepatocellular carcinoma one of the two clusters contains only $16$ patient, we show the corresponding plots as Supplementary Figure S$18$).
In all cases, the highly significant p-values prove that the information retrieved from the evolutionary models is effective in stratifying patients in well-separated risk groups. 

More in detail, brain lower grade gliomat (Fig. \ref{fig:pred}{\bf D})\\
\noindent \cite{doi:10.1056/NEJMoa1402121} is characterized by the presence of three risk groups. The low risk cluster is characterized by evolutionary trajectories comprising  an \texttt{OR} relation between mutations of \textit{IDH1} and \textit{IDH2}, which are typically associated with good prognosis \cite{doi:10.1056/NEJMoa1402121}. Notably, the high risk cluster is characterized by mutations of \textit{EGFR}, which identifies the ``glioblastoma-like" subtype with a known poor prognosis.
Consistently, in the glioblastoma multiforme (Fig. \ref{fig:pred}{\bf F}) the cluster with better overall survival is associated to the evolutionary steps comprising  mutations of \textit{TP53}, which may confirm the role of such gene as a tumor suppressor gene for this cancer type \cite{kraus2000molecular}.
In the breast invasive carcinoma (Fig. \ref{fig:pred}{\bf E}) two clusters are found and the main covariate that seems to affect the overall survival is the presence of the mutation of \textit{PIK3CA}, confirming the hypothesized positive prognostic significance of such mutation \cite{kalinsky2009pik3ca}.

The analysis of the kidney renal cell carcinoma (Fig. \ref{fig:pred}{\bf G}) returns a high number of significant covariates, which is consistent with the observed low predictability value for this cancer type.
Strikingly, the presence of the mutation of \textit{VHL} is associated to the low risk cluster, which may be in accordance to the fact that the inactivtion of such gene \textit{VHL} is a common biomarker of bad prognosis \cite{brauch2000vhl}. 
With respect to the skin cutaneous melanoma (Fig. \ref{fig:pred}{\bf H}), two different clusters are discovered, with the bad prognosis one associated to mutations on \textit{BRAF} \cite{bhatia2015impact}. 
Finally, the uterine corpus endometrial carcinoma (Fig. \ref{fig:pred}{\bf I}) shows an high number of relevant drivers, including mutations of \textit{PTEN}, \textit{TP53} and \textit{PIK3CA} \cite{doi:10.1056/NEJMoa1402121}.

Interestingly, for $3$ out of $7$ tumors, the evolutionary time $\tau$ is selected as a relevant covariate, suggesting that the tumor progression time plays an important role in the clinical outcome.
\textcolor{black}{All in all, even though high-level clinical descriptors such as the overall survival might be affected by the presence of unconsidered variables and clinical covariates, we have here shown that a relatively limited number of features of the progression models inferred by \alg{} are sufficient to stratify patients in well-divided risk groups, which in turn points at key molecular differences that might be further investigated. 
}

\paragraph{\textcolor{black}{Analysis of disjunctive relations in glioma models.}}
\textcolor{black}{Disjunctive relations may help grouping driver mutations with similar influence on phenotype, since a tumor might, for instance, exploit independent trajectories hitting the same pathway or cellular function \cite{babur2015systematic}.}
\textcolor{black}{As a proof of principle, we here focused our analysis on the disjunctive relations of the HESBCN models of gliomas that are found to be  significant covariates in the Regularized Cox Regression analysis.}

\textcolor{black}{In the case of brain lower grade glioma (Fig. \ref{fig:pred}{\bf D}), the HESBCN model includes a significant disjunctive relation, namely mutations of: \textit{TP53} \texttt{OR} \textit{IDH1} \texttt{OR} \textit{IDH2} $\triangleright$ \textit{CIC}.
Genes \textit{IDH1} and \textit{IDH2} encode for  the isocitrate dehydrogenase enzyme in different cell compartments, namely in cytosol and peroxysomes the former, in mithocondria the latter. This enzyme catalyzes the conversion of isocitrate to alpha-ketoglutarate, while reducing NADP to NADPH.  There is evidence that variations in the utilization of this reaction are associated with increased glutamine reductive carboxylation and affect redox balance, glycolysis and oxidative phosphorylation \cite{mullen2012reductive, damiani2017metabolic}. We can speculate that mutations in these genes cause similar rearrangements in cell metabolism.}
\textcolor{black}{The third gene involved in the disjunctive rule (i.e. \textit{TP53}) encodes a tumor suppressor transcription factor instead. Although recent studies have shown that \textit{TP53} has a role in the regulation of both glycolysis and oxidative phosphorylation \cite{liu2019tumor}, the encoded protein responds to diverse cellular stresses to regulate expression of target genes, thereby inducing cell cycle arrest, apoptosis, senescence, DNA repair. Hence, speculating that mutations in this gene may phenocopy mutations in \textit{IDH1} and \textit{IDH2} is a tempting, but more hazardous hypothesis.}

\textcolor{black}{Along similar lines, the glioblastoma multiforme model (Fig. \ref{fig:pred}{\bf F}) highlights the presence of the disjunction involving mutations of:
\textit{TP53} \texttt{OR} \textit{PIK3R1} \texttt{OR} \textit{PTEN} $\triangleright$ \textit{ATRX}. 
\textit{PIK3R1} is a member of the PI3K/AKT signal transduction pathway \cite{oskouian2010cancer} and plays an important role in the metabolic action of insulin and, hence, in the regulation of glycolysis \cite{asano2007role}. 
\textit{PTEN} is known to be a major antagonist of PI3K activity in the PI3K-AKT pathway \cite{vazquez2000pten} and it is also supposed to be involved in the regulation of energy metabolism in the mitochondria \cite{liu2012pten}.
\textit{PTEN} also regulates P53 protein levels and activity via distinct mechanisms \cite{freeman2003pten}, whereas the  role of mutant P53 in signaling pathways associated with glioblastoma multiforme is more elusive. 
These considerations support the hypothesis of a complex interplay involving such genes in this tumor type, which is consistent with the inferred disjunctive relation included in the HESBCN model. }








\textcolor{black}{Overall, these results show that the logical formulas inferred by \alg{} include mutations of genes possibly involved in key molecular pathways and higher-level cellular functions (e.g. energy metabolism), and which the tumor might independently exploit to further progress.}


\paragraph{\textcolor{black}{Application of \alg{} to tumors including distinct molecular subtypes.}}

\textcolor{black}{When reconstructing population-level models of tumor evolution from cross-sectional binarized mutational profiles of cancer patients, it is good practice to first stratify the samples into distinct molecular subtypes (as suggested, e.g. in \cite{caravagna2016algorithmic}), in order to reduce the possible confounding effects deriving from processing highly heterogeneous mixtures of samples.}

\textcolor{black}{However, this task may present pitfalls and an effective stratification of samples into distinct subtypes might not always be possible. Hence, in order to assess the capability of \alg{} in dissecting tumor types characterized by high heterogeneity, we performed two additional case studies.}

\textcolor{black}{We first applied \alg{} to a pan-glioma dataset obtained by merging all lower-grade glioma ($510$ samples) and glioblastoma ($338$ samples) tumours (Supplementary Figure S21). 
We then mapped the samples of the distinct subtypes onto the evolution model inferred with our approach. }
\textcolor{black}{Importantly, the HESBCN model highlights the presence of subtype-specific non-overlapping evolutionary trajectories. 
More in detail, consistently with \cite{doi:10.1056/NEJMoa1402121},   \textit{IHD1}-mutant cancers are found to be enriched for lower-grade gliomas and to be involved in two major trajectories: the first one presenting additional mutations in \textit{TP53} and \textit{ATRX} genes (CIMP subtypes \cite{CECCARELLI2016550}) and the second one involving mutations in \textit{CIC} and \textit{NOTCH1} (codel subtype \cite{CECCARELLI2016550}). 
\textit{IDH1} wild-type cancers \cite{CECCARELLI2016550} are instead mostly glioblastomas and are characterized by molecular evolutionary trajectories involing mutations in \textit{EGFR}, \textit{NF1}, \textit{PTEN} or \textit{RB1} genes.}

\textcolor{black}{We also analyzed the HESBCN model inferred from colorectal adenocarcinoma samples, by focusing on the two known molecular subtypes, i.e.  microsatellite stable (MSS, $396$ samples) and microsatellite instable (MSI, $62$ samples). The HESBCN model associates canonical colorectal cancer drivers such as mutations of \textit{APC}, \textit{TP53} and \textit{KRAS} to MSS cancers, while MSI tumours show a broad range of driver mutations comprising molecular trajectories involving \textit{DMD}, \textit{SPTA1}, \textit{FBXW7} and \textit{KMT2D} (see the Supplementary Figure S22).
Collectively, these results demonstrate the effectiveness of \alg{} in representing different molecular subtypes within a unique evolution model. }

\section{Discussion}

The possibility of exploiting sequencing data to reliably predict the likely clinical outcome of a given cancer patient, and possibly intervene to halt or slow down the disease progression, may have an important impact on downstream clinical practices and therapeutic strategies. 

In this regard, we have shown that cancer progression models, even when inferred from low-resolution (binarized) mutational profiles of cross-sectional samples, may deliver accurate predictions on patients' survival. 
Accordingly, targeted sequencing of specific gene panels might be a viable and cost-effective strategy to position a given patient onto the expected cancer progression route at diagnosis time, possibly anticipating the next evolutionary steps. 

\textcolor{black}{Clearly, the problem is far from being solved. From the computational perspective, one limitation is that structural learning of Bayesian networks is an extremely hard task and there is no guarantee of converging or reaching global optima via MCMC search schemes \cite{ramazzotti2018modeling,ramazzotti2019efficient}.}
 
\textcolor{black}{
We also note that important results on the inference of the temporal ordering of genetic lesions in single tumours were  achieved by estimating cancer timelines from Variant Allele Frequency profiles of single patients, via league model analysis and subclonal deconvolution \cite{gerstung2020evolutionary}. 
However, as \alg{} pools together data from multiple patients and drops Variant Allele Frequency information, it cannot -- by construction -- shed any light on the temporal evolution sequence that holds for a specific patient. Instead, \alg{} allows one to infer a statistically robust population-level estimator of cancer evolution. Ideas from these two complementary approaches might lead to a more comprehensive characterization of tumours timelines, which we leave for future works.}

\textcolor{black}{A further pitfall of approaches processing low-resolution (binarized) data from bulk samples is} due to the impact of intra-tumor heterogeneity, which is a major cause of therapy failure and relapse \cite{sottoriva2013intratumor} and which can undermine the accuracy of any population model. 
As a consequence, many new methods attempt to deliver cancer evolution models at the resolution of the single tumor \cite{jahn2016tree,ramazzotti2017learning,zafar2019siclonefit}, taking advantage of the recent advances in single-cell DNA sequencing techniques and, more recently, of the opportunity of calling variants from RNA-sequencing data \cite{moraveccancer}, despite the typically high levels of technical and biological noise \cite{patruno2020review}.

In this respect, it would be \textcolor{black}{interesting} to investigate how to combine single-tumor models within comprehensive predictive population models as here proposed, for instance by employing transfer learning to find patterns of repeated evolution \cite{caravagna2018detecting}. In addition, the rise of longitudinal sequencing experiments, e.g. from patient-derived organoids, may allow one to assess the impact of selected therapeutic strategies on the predicted evolution \cite{ramazzotti2020longitudinal}, with important translational repercussions. 

%
%


\section*{Acknowledgements}

We thank Marco Antoniotti, Lucrezia Patruno, Francesco Craighero, Davide Maspero and Gianluca Ascolani for useful discussions. 
We also thank Pablo Herrera Nieto and Ramon Diaz-Uriarte for their valuable comments on the first version of the manuscript.

\section*{Funding}

This work was partially supported by a Bicocca 2020 Starting Grant to FA and DR. DR was also supported by a Premio Giovani Talenti of the University of Milan-Bicocca. This work was also supported by the CRUK/AIRC Accelerator Award \#22790
``Single-cell Cancer Evolution in the Clinic''. Financial support from the Italian Ministry of University and Research (MIUR) through the grant ‘Dipartimenti di Eccellenza 2017’ to the Department of Biotechnology and Bioscences of University of Milan-Bicocca is also acknowledged.

\bibliographystyle{abbrv}
\bibliography{references.bib}

\end{document}